\documentclass[12pt]{article}
\usepackage[colorlinks=true, urlcolor=red, linkcolor=blue, citecolor=blue]{hyperref}
\usepackage{times,amsmath,amsfonts,amsthm,amssymb,eucal}
\usepackage{graphicx,ifthen}
\usepackage{color}
\usepackage{mathrsfs}
\usepackage{bm}

\usepackage{natbib}

\usepackage{tabularx}
\usepackage{array}
\usepackage{ragged2e}
\usepackage{lscape}
\usepackage{multirow}
\usepackage{booktabs}

\usepackage{caption}
\usepackage{subcaption}
 
\newcommand{\blind}{0}

\DeclareGraphicsExtensions{.gif,.pdf,.png,.jpg,.tiff}

\def\({\left(}
\def\){\right)}
\def\[{\left[}
\def\]{\right]}
\def\a{\alpha}\def\b{\beta}

\def\s{\sigma}

\def\RSF{\mathscr}

 \def\gg{{\RSF G}}

  1
 1
 1

\newdimen\bigindent
\newdimen\smallindent
\def\quoteindent{\advance\leftskip by\bigindent\advance\rightskip
                 by\bigindent}
\newskip\proclaimskipamount
\def\proclaimskip{%
  \par\ifdim\lastskip<\proclaimskipamount
  \removelastskip\vskip\proclaimskipamount\fi}

\def\Demo#1{\par\ifdim\lastskip<\proclaimskipamount
            \removelastskip\proclaimskip\fi
            \noindent\sl#1. \hskip\smallindent\rm}

\def\DemoSection#1{\par\ifdim\lastskip<\proclaimskipamount
             \removelastskip\proclaimskip\fi
             #1\hskip\smallindent\rm}

\def\Quote{\begin{quotation}\normalfont\small}
\def\EndQuote{\end{quotation}\rm}

\def\bct{\begin{center}}
\def\ect{\end{center}}
\def\Array{\begin{eqnarray*}}
\def\EndArray{\end{eqnarray*}}
\def\Enumerate{\begin{enumerate}}
\def\EndEnumerate{\end{enumerate}}
\def\Itemize{\begin{itemize}}
\def\EndItemize{\end{itemize}}
\def\Eq{\begin{equation}}
\def\EndEq{\end{equation}}
\def\EqArray{\begin{eqnarray}}
\def\EndEqArray{\end{eqnarray}}
\def\mref#1{(\ref{#1})}

\def\Tabular{\begin{tabular}}
\def\EndTabular{\end{tabular}}
\def\FlushLeft{\begin{flushleft}}
\def\EndFlushLeft{\end{flushleft}}
\def\FlushRight{\begin{flushright}}
\def\EndFlushRight{\end{flushright}}

\def\pmb#1{\setbox0=\hbox{#1}%
 \kern-0.010em\copy0\kern-\wd0
 \kern0.035em\copy0\kern-\wd0
 \kern0.010em\raise.0233em\box0}

\def\bart{\text{BART}}
\def\biv{\mathrm{bivariate}}
\def\bias{{\text{Bias}}}
\def\CF{\text{CF}}

\def\E{\mathrm{E}}

\def\eps{\varepsilon}

\def\honest{\mathrm{honestRF}}

\def\N{{\text N}
\def\MSE{{\text MSE}}}

\def\CF{\text{CF}}

\def\SCF{\text{synCF}}

\def\tauhat{\hat{\tau}}

\def\VT{\text{VT}}
\def\VTI{\text{VT-I}}

\def\X{{\bf X}}
\def\x{{\bf x}}
\def\Y{{\bf Y}}

\def\Yhat{{\hat Y}}

\def\EE{\mathbb{E}}

\def\PP{\mathbb{P}}

\newcommand{\rcode}[1]{{\ttfamily #1}}

\newcolumntype{P}[1]{>{\RaggedRight\hspace{0pt}}p{#1}}
\newcolumntype{L}[1]{>{\raggedright\let\newline\\\arraybackslash\hspace{0pt}}m{#1}}
\newcolumntype{C}[1]{>{\centering\let\newline\\\arraybackslash\hspace{0pt}}m{#1}}
\newcolumntype{R}[1]{>{\raggedleft\let\newline\\\arraybackslash\hspace{0pt}}m{#1}}

\begin{document}

\def\spacingset#1{\renewcommand{\baselinestretch}%
{#1}\small\normalsize} \spacingset{1}


  \title{\bf Estimating Individual Treatment Effect \\ in
  Observational Data Using Random Forest Methods}


  \author{
    Min Lu\\
    Division of Biostatistics, University of Miami\vspace{5pt}\\
    Saad Sadiq\\
    Department of Electrical and Computer Engineering, University of Miami\vspace{5pt}\\
    Daniel J. Feaster\\
    Division of Biostatistics, University of Miami\vspace{5pt}\\
    and \\
    Hemant Ishwaran\\
    Division of Biostatistics, University of Miami\vspace{10pt}\\
  }

  \maketitle
  
  \bigskip
  \noindent
  Estimation of individual treatment effect in observational data is
  complicated due to the challenges of confounding and selection bias.
  A useful inferential framework to address this is the counterfactual
  (potential outcomes) model which takes the hypothetical stance of
  asking what if an individual had received {\it both} treatments.  Making
  use of random forests (RF) within the counterfactual framework we
  estimate individual treatment effects by directly modeling the
  response.  We find accurate estimation of individual treatment
  effects is possible even in complex heterogeneous settings but that
  the type of RF approach plays an important role in accuracy.
  Methods designed to be adaptive to confounding, when used in
  parallel with out-of-sample estimation, do best.  One method found
  to be especially promising is counterfactual synthetic forests.  We
  illustrate this new methodology by applying it to a large
  comparative effectiveness trial, Project Aware, in order to explore
  the role drug use plays in sexual risk.  The analysis reveals
  important connections between risky behavior, drug usage, and sexual
  risk.

  \bigskip

\noindent%
\vfill

\newpage
\spacingset{1.45} 

\section{Introduction}

Even for a medical discipline steeped in a tradition of randomized
trials, the evidence basis for only a few guidelines is based on
randomized trials~\citep{tricoci2009scientific}. In part this is due
to continued development of treatments, in part to enormous expense of
clinical trials, and in large part to the hundreds of treatments and
their nuances involved in real-world, heterogeneous clinical
practice. Thus, many therapeutic decisions are based on observational
studies. However, comparative treatment effectiveness studies of
observational data suffer from two major problems: only partial
overlap of treatments and selection bias. Each treatment is to a
degree bounded within constraints of indication and
appropriateness. Thus, transplantation is constrained by variables
such as age, a mitral valve procedure is constrained by presence of
mitral valve regurgitation. However, these boundaries overlap widely,
and the same patient may be treated differently by different
physicians or different hospitals, often without explicit or evident
reasons. Thus, a fundamental hurdle in observational studies
evaluating comparative effectiveness of treatment options is to
address the resulting selection bias or confounding. Naively
evaluating differences in outcomes without doing so leads to biased
results and flawed scientific conclusions.

Formally, let $\{(\X_1,T_1,Y_1),\ldots,(\X_n,T_n,Y_n)\}$ denote the
data where $\X_i$ is the covariate vector for individual $i$ and $Y_i$
is the observed outcome.  Here $T_i$ denotes the treatment group of
$i$.  For concreteness, let us say $T_i=0$ represents the control
group, and $T_i=1$ the intervention group.  Our goal is to estimate
the {\it individual treatment effect} (ITE), defined as the difference in
the mean outcome for an individual under both treatments, conditional
on the observed covariates.  More formally, let $Y_i(0)$ amd $Y_i(1)$
denote the potential outcome for $i$ under treatments $T_i=0$ and
$T_i=1$, respectively.  Given $\X_i=\x$, the ITE for $i$ is defined as
the conditional mean difference in potential outcomes
\Eq 
\tau(\x) =
\EE\[Y_i(1)|\X_i=\x\] - \EE\[Y_i(0)|\X_i=\x\] .
\label{ite}
\EndEq 
Definition~\mref{ite} relies on what is called the
counterfactual framework, or potential outcomes model
\citep{splawa1990application, rubin1974estimating}.  In this
framework, one plays the game of hypothesizing what would have
happened if an individual $i$ had received both treatments.  However,
the difficulty with estimating~\mref{ite} is that although 
potential outcomes $\{Y_i(0), Y_i(1)\}$ are hypothesized to exist,
only the outcome $Y_i$ from the actual treatment assignment is
observed.  Without additional assumptions, it is not possible in
general to estimate~\mref{ite}.  A widely used assumption to resolve
this problem, and one that we adopt, is the assumption of strongly
ignorable treatment assignment (SITA).  This assumes that treatment
assignment is conditionally independent of the potential outcomes
given the variables; i.e., $T \perp \{Y(0),Y(1)\} \, \, |\,\, \X$.
Under the assumption of SITA, we have
\EqArray
\tau(\x) 
&=& \EE\[Y(1)|T=1,\X=\x\] - \EE\[Y(0)|T=0,\X=\x\] \nonumber\\
&=& \EE\[Y|T=1,\X=\x\] - \EE\[Y|T=0,\X=\x\] .
\label{ite.sita}
\EndEqArray Thus, $\tau(\x)$ becomes estimable under SITA as it can be
expressed in terms of conditional expectations of observable values.
It should be emphasized that without SITA one cannot guarantee
estimability of $\tau(\x)$ because $\EE\[Y|T=j,\X=\x\] =
\EE\[Y(j)|\X=\x\]$ does not hold 
in general.  SITA also provides a
means for estimating the average treatment effect (ATE), a standard
measure of performance in non-heterogeneous treatment settings.  The
ATE is defined as $\tau_0 = \EE\[Y_i(1)\] - \EE\[Y_i(0)\] =
\EE\[\tau(\X)\]$.  By averaging over the distribution of $\X$
in~\mref{ite.sita}, 
\EqArray 
\tau_0 &=& \EE\biggl\{\EE\[Y|T=1,\X=\x\] - \EE\[Y|T=0,\X=\x\] \biggr\}
\nonumber\\
&=& \EE\[Y|T=1\] - \EE\[Y|T=0\].
\label{ate.sita}
\EndEqArray
Thus, SITA also ensures that $\tau_0$ is estimable.  

Although direct estimation of~\mref{ite.sita} or~\mref{ate.sita} is
possible by using mean treatment differences in cells with the same
$\X$, due to the curse of dimensionality this method only
works when $\X$ is low dimensional.  Propensity score
analysis~\citep{rosenbaum1983central} is one means to overcome this
problem.  The propensity score is defined as the conditional
probability of receiving the intervention given $\X=\x$, denoted here
by $e(\x) = \PP\{T=1|\X=\x\}$.  Under the assumption of SITA, the
propensity score possesses the so-called balancing property.  This
means that $T$ and $\X$ are conditionally independent given $e(\X)$.
Thus variables $\X$ are balanced between the two treatment groups
after conditioning on the propensity score, thereby approximating a randomized
clinical trial~\citep{rubin2007design}.  Importantly, the propensity
score is the coarsest possible balancing score, thus not only does it
balance the data, but it does so by using the coarsest possible
conditioning, thus helping to mitigate the curse of dimensionality.
In order to use the propensity score for treatment effect estimation,
\cite{rosenbaum1983central} further showed that if the propensity
score is bounded $0<e(\X)<1$ and SITA holds, then treatment assignment
is conditionally independent of the potential outcomes given the
propensity score; i.e., $T \perp \{Y(0),Y(1)\} \,\, |\,\, e(\X)$.
This result is the foundation for ATE estimators based on
stratification or matching of the data on propensity scores.  However, this is not
the only means for using the propensity score to estimate treatment
effect.  Another approach is to use SITA to derive
weighted estimators for the ATE.  Analogous to~\mref{ite.sita}, under
SITA one has
$$
\EE\[\frac{T Y}{e(\X)}\biggl|\X=\x\]= \EE\[Y|T=1,\X=\x\],\hskip15pt
\EE\[\frac{(1-T) Y}{1-e(\X)}\biggl|\X=\x\]= \EE\[Y|T=0,\X=\x\],
$$ 
which is the basis for ATE weighted propensity score estimators.
See for example, ~\cite{hirano2003efficient}
and~\cite{lunceford2004stratification}.

\subsection{Individual treatment effect estimation}

As mentioned, our focus is on estimating the ITE.  Although
effectiveness of treatment in observational studies has traditionally
been measured by the ATE, the practice of individualized medicine,
coupled with the increasing complexity of modern studies, have shifted
recent efforts towards a more patient-centric
view~\citep{lamont2015identification}.  Accommodating complex
individual characteristics in this new landscape has proven
challenging, and for this reason there has been much interest in
leveraging cutting-edge approaches, especially those from machine
learning.  Machine learning techniques such as random
forests~\citep{breiman2001random} (RF) provide a principled approach
to explore a large number of predictors and identify replicable sets
of predictive factors.  In recent innovations these RF approaches have
been used specifically to uncover subgroups with differential
treatment responses~\citep{su2009subgroup,
  su2011interaction,foster2011subgroup}.  Some of these, such as the
virtual twins approach~\citep{foster2011subgroup}, build on the idea
of counterfactuals.  Virtual twins uses RF as a first step to create
separate predictions of outcomes under both treatment and control
conditions for each trial participant by estimating the counterfactual
treatment outcome.  In the second step, tree-based predictors are used
to uncover variables that explain differences in the person-specific
treatment and the characteristics associated with subgroups.  In a
different approach,~\cite{wager2015estimation} describe causal forests
for ITE estimation.  Others have sought to use RF as a first step in
propensity score analysis as a means to nonparametrically estimate the
propensity score.  \cite{lee2010improving} found that RF estimated
propensity scores resulted in better balance and bias reduction than
classical logistic regression estimation of propensity scores. 

In this manuscript, we look at several different RF methods for
estimating the ITE.  A common thread among these methods is that they
all directly estimate the ITE, and each does so
without making use of the propensity score.  Although propensity score
analyses have traditionally been used for estimating the ATE,
non-ATE estimation generally takes a more direct approach
by modeling 
the outcome.  Typically this is done by using some form of regression
modeling.  For example, this is the key idea underlying the widely
used ``g-formula'' algorithm~\citep{robins1999estimation}.  Another
example are Bayesian tree methods for regression surface modeling
which have been successfully used to identify causal
effects~\citep{hill2011bayesian}.  The basis for all of these
approaches rests on the assumption of SITA.  Assuming that
the outcome $Y$ satisfies
$$
Y=f(\X,T)+\eps,
$$
where $\E(\eps)=0$ and $f$ is the unknown regression function, and assuming
that SITA holds, we have
\EqArray
\tau(\x) 
&=& \EE\[Y(1)|T=1,\X=\x\] - \EE\[Y(0)|T=0,\X=\x\] \nonumber\\
&=& \EE\[Y|T=1,\X=\x\] - \EE\[Y|T=0,\X=\x\] \nonumber\\
&=& f(\x,1)-f(\x,0).
\label{regression.validation}
\EndEqArray 
Therefore by modeling 
$f(\x,T)$, we obtain a means for directly estimating the ITE.

Our proposed RF methods for direct estimation of the ITE are described
in Section 2.  In Section 3, we use three sets of challenging
simulations to assess performance of these methods.  We find those
methods with greatest adaptivity to potential confounding, when
combined with out-of-sample estimation, do best.  One particularly
promising approach is a counterfactual approach in which separate
forests are constructed using data from each treatment assignment. To
estimate the ITE, each individual's predicted outcome is obtained from
their treatment assigned forest.  Next, the individual's treatment is
replaced with the counterfactual treatment and used to obtain the
counterfactual predicted outcome from the counterfactual forest; the
two values are differenced to obtain the estimated ITE.  This is an
extension of the virtual twin approach, modified to allow for greater
adaptation to potentially complex treatment responses across
individuals.  Furthermore, when combined with synthetic
forests~\citep{ishwaran2014synthetic}, performance of the method is
further enhanced due to reduced bias.  In Section 4, we illustrate the
new counterfactual synthetic method on a large comparative
effectiveness trial, Project Aware~\citep{metsch2013effect}.  An
original goal of the project was to determine if risk reduction
counseling for HIV negative individuals at the time of an HIV test had
an impact on cumulative incidence of sexually transmitted infection
(STI). However, secondary outcomes included continuous and count
outcomes such as total number of condomless sexual episodes, number of
partners, and number of unprotected sex acts.  This trial had a
significant subgroup effect in which men who have sex with men showed
a surprising higher rate of STI when receiving risk-reduction
counseling.  This subgroup effect makes this trial ideal for looking
at heterogeneity of treatment effects in randomized studies.  The
trial also had significant heterogeneity by drug use. In particular,
substance use is associated with higher rates of HIV testing, and
Black women showing the highest rates of HIV testing in substance
use treatment.  Drug use is a modifiable exposure (treatment)
variable and we can therefore use our methods for addressing
heterogeneity in observational data to study its impact on sexual
risk.  As detailed in Section 4, we show how our methods can be used
to examine whether observed drug use differences are merely proxies
for differences on other observed variables.

\section{Methods for estimating individual treatment effects}

Here we describe our proposed RF methods for estimating the
ITE~\mref{ite}.  Also considered are two comparison methods.  
The methods considered in this paper are as
follows:
\Enumerate\setlength\itemsep{-10pt}
\item[1.]
Virtual twins (VT).
\item[2.]
Virtual twins interaction (VT-I).
\item[3.]
Counterfactual RF (CF).
\item[4.]
Counterfactual synthetic RF (synCF).
\item[5.]
Bivariate RF (bivariate).
\item[6.]
Honest RF (honest RF).
\item[7.]
Bayesian Adaptive Regression Trees (BART).
\EndEnumerate
Virtual twins is the original method proposed
by~\cite{foster2011subgroup} mentioned earlier.  We also consider an
extension of the method, called virtual twins interaction, which
includes forced interactions in the design matrix for more adaptivity.
Forcing treatment interactions for adaptivity may have a limited
ceiling, which is why we propose the counterfactual RF method.  In
this method we dispense with interactions and instead fit separate
forests to each of the treatment groups.  Counterfactual synthetic RF
uses this same idea, but uses synthetic forests in place of Breiman
forests, which is expected to further improve adaptivity.  Thus, this
method, and the previous RF methods, are all proposed enhancements to
the original virtual twins method.  All of these share the common
feature that they provide a direct estimate for the ITE by estimating
the regression surface of the outcome.  This is in contrast to our
other proposed procedure, bivariate RF, which takes a missing data
approach to the problem.  There has been much interest in the
literature in viewing causual effect analysis as a missing data
problem~\citep{ghosh2015penalized}. Thus, we propose here a novel
bivariate imputation approach using RF.  Finally, the last two
methods, honest RF and BART, are included as comparison procedures.
Like our proposed RF methods, they also directly estimate the ITE.
Note that while BART~\citep{chipman2010bart} is a tree-based method,
it is not actually a RF method.  We include it, however, because of
its reported success in applications to causal
inference~\citep{hill2011bayesian}.  In the following sections we
provide more details about each of the above methods.

\subsection{Virtual twins}

\cite{foster2011subgroup} proposed Virtual Twins (VT)
for estimating counterfactual outcomes.  In this approach, RF
is used to regress $Y_i$ against $(\X_i,T_i)$.
To obtain a counterfactual estimate for an individual $i$,
one creates a VT data point, similar in all regards to the original
data point $(\X_i,T_i)$ for $i$, but with the observed treatment $T_i$
replaced with the counterfactual treatment $1-T_i$.  Given an
individual $i$ with $T_i=1$, one obtains the RF predicted value
$\Yhat_i(1)$ by running $i$'s unaltered data down the forest.  To
obtain $i$'s counterfactual estimate, one runs the altered
$(\X_i,1-T_i)=(\X_i,0)$ down the forest to obtain the counterfactual estimate
$\Yhat_i(0)$.  The counterfactual ITE estimate is defined
as $\Yhat_i(1)-\Yhat_i(0)$.  A similar argument is applied when
$T_i=0$.  If $\Yhat_{\VT}(\x,T)$ denotes the predicted value for
$(\x,T)$ from the VT forest,  the VT counterfactual estimate for
$\tau(\x)$ is
$$
\tauhat_\VT(\x) = \Yhat_{\VT}(\x,1)-\Yhat_{\VT}(\x,0).
$$ 
As noted in~\cite{foster2011subgroup}, the VT
approach can be improved by manually including treatment interactions
in the design matrix.  Thus, one runs a RF regression with $Y_i$
regressed against $\(\X_i,T_i,\X_i T_i\)$.  The inclusion of the
pairwise interactions $\X_iT_i$ is not conceptually necessary for VT,
but~\cite{foster2011subgroup} found in numerical work that it
improved results.  We write $\tauhat_\VTI(\x)$ to
denote the ITE estimate under this modified VT interaction
model.

There is an important computational point that we mention here that
applies not only to the above procedure, but also to many of the
proposed RF methods.  That is when implementing a RF procedure, we
attempt to use out-of-bag (OOB) estimates whenever possible.  This is
because OOB estimates are generally much more accurate 
than insample (inbag) estimates~\citep{breiman1996oob}.  Because
inbag/OOB estimation is not made very clear in the RF literature, it
is worth discussing this point here as readers may be unaware of this
important distinction.  OOB refers to out-of-sample (cross-validated)
estimates.  Each tree in a forest is constructed from a bootstrap
sample which uses approximately 63\% of the data.  The remaining 37\%
of the data is called OOB and are used to calculate an OOB predicted
value for a case.  The OOB predicted value is defined as the predicted
value for a case using only those trees where the case is OOB. For
example, if 1000 trees are grown, approximately 370 will be used in
calculating the OOB estimate for the case.  The inbag predicted value,
on the other hand, uses all 1000 trees.

 To illustrate how OOB estimation applies to
VT, suppose that case $\x$ is assigned treatment $T=1$.  Let
$\Yhat_{\VT}^*(\x,T)$ denote the OOB predicted value for $(\x,T)$.
The OOB counterfactual estimate for $\tau(\x)$ is
$$
\tauhat_\VT(\x) = \Yhat_{\VT}^*(\x,1)-\Yhat_{\VT}(\x,0).
$$ 
Note that $\Yhat_{\VT}(\x,0)$ is not OOB.  This is because $(\x,0)$ is
a new data point and technically speaking cannot have an OOB predicted
value as the observation is not even in the training data.  In a
likewise fashion, if $\x$ were assigned treatment $T=0$, the OOB
estimate is
$$
\tauhat_\VT(\x) = \Yhat_{\VT}(\x,1)-\Yhat_{\VT}^*(\x,0).
$$ 
OOB counterfactual estimates for  VT interactions,
$\tauhat_\VTI(\x)$, are defined
analogously.

\subsection{Counterfactual RF}

As mentioned earlier, adding treatment interactions to the design
matrix may have a limited ceiling for adaptivity and thus we introduce
the following important extension to $\tauhat_\VTI$.  Rather than
fitting a single forest with forced treatment interactions, we instead
fit a separate forest to each treatment group to allow for greater
adaptivity.  This modification to VT was mentioned briefly in the
paper by~\cite{foster2011subgroup} although not implemented.  A
related idea was used by~\cite{dasgupta2014risk} to estimate
conditional odds ratios by fitting separate RF to different exposure
groups.

In this method, forests $\CF_1$ and $\CF_0$ are fit separately to data
$\{(\X_i,Y_i):T_i=1\}$ and $\{(\X_i,Y_i):T_i=0\}$, respectively. To
obtain a counterfactual ITE estimate, each data point is run down its
natural forest, as well as its counterfactual forest.  If
$\Yhat_{\CF,j}(\x,T)$ denotes the predicted value for $(\x,T)$ from
$\CF_j$, for $j=0,1$, the counterfactual ITE estimate is
$$
\tauhat_\CF(\x) = \Yhat_{\CF,1}(\x,1)-\Yhat_{\CF,0}(\x,0).
$$ 
We note that just as with VT estimates, OOB values are utilized
whenever possible to improve stability of estimated values.
Thus, if $\x$ is assigned treatment $T=1$, the OOB
ITE estimate is
$$
\tauhat_\CF(\x) = \Yhat_{\CF,1}^*(\x,1)-\Yhat_{\CF,0}(\x,0),
$$ 
where $\Yhat_{\CF,1}^*(\x,1)$ is the OOB predicted value for $(\x,1)$.
Likewise, if $\x$ is assigned treatment $T=0$, the OOB estimate is
$$
\tauhat_\CF(\x) = \Yhat_{\CF,1}(\x,1)-\Yhat_{\CF,0}^*(\x,0).
$$

\subsection{Counterfactual synthetic RF}

In a modification to the above approach, we replace Breiman RF
regression used for predicting $\Yhat_{\CF,j}(\x,T)$ with
synthetic forest regression using synthetic
forests~\citep{ishwaran2014synthetic}.  The latter are a new type of
forest designed to improve prediction performance of RF.  Using a
collection of Breiman forests (called base learners) grown under
different tuning parameters, each generating a predicted value called
a synthetic feature, a synthetic forest is defined as a secondary
forest calculated using the new input synthetic features, along with
all the original features.  Typically, the base learners used by
synthetic forests are Breiman forests grown under different nodesize
and mtry parameters.  The latter are tuning parameters used in
building a Breiman forest.  In RF, prior to splitting a tree node, a
random subset of mtry variables are chosen from the original
variables.  Only these randomly selected variables are used for
splitting the node.  Splitting is applied recursively and the tree
grown as deeply as possible while maintaining a sample size condition
that each terminal node contains a minimum of nodesize cases.  The two
tuning parameters mtry and nodesize are fundamental to the performance
of RF.  Synthetic forests exploits this and uses RF base learners
grown under different mtry and nodesize parameter values.  To
distinguish the proposed synthetic forest method from the
counterfactual approach described above, we use the abbreviation synCF
and denote its ITE estimate by $\tauhat_\SCF(\x)$:
$$
\tauhat_\SCF(\x) = \Yhat_{\SCF,1}(\x,1)-\Yhat_{\SCF,0}(\x,0),
$$ 
where $\Yhat_{\SCF,j}(\x,T)$ denotes the predicted value for
$(\x,T)$ from the synthetic RF grown using data
$\{(\X_i,T_i,Y_i):T_i=j\}$ for $j=0,1$.  As before, OOB estimation is
used whenever possible.  In particular, bootstrap samples are held
fixed throughout when constructing synthetic features and the
synthetic forest calculated from these features.  This is done to
ensure a coherent definition of being out-of-sample.

\subsection{Bivariate imputation approach}

We also introduce a new bivariate approach making use of bivariate RF
counterfactuals.  For each individual $i$, we assume the existence of
bivariate outcomes under the two treatment groups.  One of these is
the observed $Y_i$ under the assigned treatment $T_i$, the other is
the unobserved $Y_i$ under the counterfactual treatment $1-T_i$.  This
latter value is assumed to be missing.  To impute these missing
outcomes a bivariate splitting rule is used~\citep{rsf,tang2015}.  In
the first iteration of the algorithm, the bivariate splitting rule
only uses the observed $Y_i$ value when splitting a tree node.  At the
completion of the forest, the missing $Y_i$ values are imputed by
averaging OOB terminal node $Y$ values.  This results in a data set
without missing values, which is then used as the input to another
bivariate RF regression.  In this, the bivariate splitting rule is
applied to the bivariate response values (no missing values are
present at this point).  At the completion of the forest, $Y$ values
that were originally missing are reset to missing and imputed using
mean terminal node $Y$ values.  This results in a data set without missing
values, which is used once again as the input to another bivariate RF
regression.  This process is repeated a fixed number of times.  At the
completion of the algorithm, we have complete bivariate $\Y_i$
responses for each $i$, which we denote by
$\hat{\Y}_{\biv,i}=(\Yhat_{\biv,1},\Yhat_{\biv,0})$.  Note that one of
these values is the original observed (non-missing) response.
The complete bivariate values are used to
calculate the bivariate counterfactual estimate
$$
\tauhat_\biv(\x) = \Yhat_{\biv,1}(\x)-\Yhat_{\biv,0}(\x).
$$  

\subsection{Honest RF}

As a comparison procedure, 
we consider the honest RF method described in Procedure 1
of~\cite{wager2015estimation}.  In this method, a RF is run by
regressing $Y_i$ on $(\X_i,T_i)$, but using only a randomly selected
50\% subset of the data.  When fitting RF to this training data, a
modified regression splitting rule is used.  Rather than splitting
tree nodes by maximizing the node variance, honest RF instead uses a
splitting rule which maximizes the treatment difference within a node
\citep[see Procedure~1 and Remark~1 in][]{wager2015estimation}. Once
the forest is grown, the terminal nodes of the training forest are
repopulated by replacing the training $Y$ with the $Y$ values from the
data that was held out.  The purpose of this hold out data is to
provide honest estimates and is akin to the role played by the OOB
data used in our previous procedures.  The difference between the hold
out $Y$ values under the two treatment groups is determined for each
terminal node and averaged over the forest.  This forest averaged
value represents the honest forest ITE estimate.  We denote the honest
RF estimate by $\tauhat_\honest(\x)$.

\subsection{BART}

\cite{hill2011bayesian} described a causal inferential approach based
on BART~\citep{chipman2010bart}.  The BART procedure is a type of
ensembled backfitting algorithm based on Bayesian regularized tree
learners. Because the algorithm repeatedly refits tree residuals, BART
can be intuitively thought of as a Bayesian regularized tree boosting
procedure. \cite{hill2011bayesian} proposed using BART to directly
model the regression surface to estimate potential outcomes.
Therefore, this is similar to VT, but where RF is replaced with BART.
The BART ITE estimate is defined as
$$
\tauhat_\bart(\x) = \Yhat_{\bart}(\x,1)-\Yhat_{\bart}(\x,0)
$$ 
where $\Yhat_{\bart}(\x,T)$ denotes the predicted value for
$(\x,T)$ from BART.  Note that due to the highly adaptive nature of
BART, no forced interactions are included in the design matrix.

\section{Simulation experiments}

Simulation models with differing types of heterogeneous
treatment effects were used to assess performance of the different
methods. We simulated $p=20$ independent covariates, where covariates
$X_1,\ldots,X_{11}$ were drawn from a standard $\N(0,1)$, and
covariates $X_{12},\ldots,X_{20}$ from a Bernoulli$(0.5)$.  Three
different models were used for the outcome $Y$, while a common
simulation model was used for the treatment variable $T\in\{0,1\}$.
For the latter, a logistic regression model was used to simulate $T$
in which the linear predictor $F(\X)$ defined on the logit scale was
$$
F(\X) = 
-2 + .028   X_1 - .374   X_2 - .03   X_3 + .118   X_4 - 0.394   X_{11} +
0.875   X_{12} +0.9  X_{13}.
$$
For the three outcome models, the outcome was assumed to be
$Y_i=f_j(\X_i,T_i)+\eps_i$, where $\eps_i$ were
independent $\N(0,\s^2)$.   The mean functions $f_j$ for the three
simulations were
\Array
f_1(\X,T) 
&=&   2.455 
- 1_{\{T=0\}}\times (.4  X_1 + .154  X_2 - .152  X_{11} - .126  X_{12})
- 1_{\{T=1,\,g(\X)>0\}}\\
f_2(\X,T) 
&=&   2.455 
- 1_{\{T=0\}}\times \sin(.4  X_1 + .154  X_2 - .152  X_{11} - .126  X_{12})
- 1_{\{T=1,\, g(\X)>0\}}\\
f_3(\X,T) 
&=&   2.455 
- 1_{\{T=0\}}\times \sin(.4  X_1 + .154  X_2 - .152  X_{11} - .126  X_{12})
- 1_{\{T=1,\,h(\X)>0\}}
\EndArray
where $g(\X) = .254 X_2 ^ 2 - .152 X_{11} - .4 X_{11}^2 - .126 X_{12}$
and $h(\X) = .254 X_3 ^ 2 - .152 X_{4} - .126 X_{5} - .4
X_{5}^2$. Therefore in all three models, $X_{1},X_{2},X_{11}, X_{12}$
were confounding variables, meaning that they were related to both the
treatment and the outcome variable.  In model 3, variables $X_3$ and
$X_4$ are additionally confounded.  Also, because all three models
contain treatment-covariate interactions, all models simulate scenarios of
confounded heterogeneous treatment effect (CHTE).  The type of CHTE simulated
is different for each model.  In model
1, there is a non-linear effect for treatment $T=1$.  For model 2,
there are non-linear effects for both treatment groups, and in model
3, non-linear effects are present for both treatment groups, and there
is non-overlap in covariates across treatment groups.

\subsection{Experimental settings and parameters}

The three simulation models were run under two settings for the sample
size, $n = 500$ and $n=5000$.  All simulations used $\s=0.1$ for the
standard deviation of the measurement errors.  The smaller sample size
experiments $n=500$ were repeated independently $B=1000$ times, the
larger $n=5000$ experiments were repeated $B=250$ times.  All random
forests were based on 1000 trees with mtry $=p/3$ and a nodesize of 3
with the exception of bivariate RF and synthetic RF.  For bivariate
forests, a nodesize of 1 was used \citep[following the strategy
  recommended by][]{tang2015}, while for for synthetic RF, the RF base
learners were constructed using all possible combinations of nodesize
values 1--10, 20, 30, 50, 100 and mtry values 1, 10 and 20 (for a
total of 42 forest base learners).  The bivariate procedure was
iterated 5 times (i.e., each run used a 5 step iteration procedure).
All forest computations except for honest RF were implemented using
the {\ttfamily randomForestSRC} R-package~\citep{rfsrc} (hereafter
abbreviated as RF-SRC).  The RF-SRC package implements all forms of RF
data imputation, fits synthetic forests, multivariate forests, and
utilizes openMP parallel processing for rapid computations.  For
honest RF, we used the R-package \rcode{causalForest} available at
github~\url{https://github.com/swager/causalForest}.  The nodesize was
set to 1 and 1000 trees were used.  For BART, we used the \rcode{bart}
function from the R-package \rcode{BayesTree}~\citep{bart}.  A total
of 1000 trees were used.

\subsection{Performance measures}

Performance was assessed by bias and root mean squared error (RMSE).
When calculating these measures we conditioned on the propensity
score, $e(\x)$.  This was done to assess how well a procedure could
recover treatment heterogeneity effects and to provide insight into
its sensitivity to treatment assignment.  A robust procedure should
perform well not only in regions of the data where $e(\x)=0.5$, and
treatment assignment is balanced, but also in those regions where
treatment assignment is unbalanced, $0<e(\x)<.5$ and $1>e(\x)>.5$.
Assume the data is stratified into groups $\gg=\{\gg_1,\ldots,\gg_M\}$
based on quantiles $q_1,\ldots,q_M$ of $e(\x)$.  Given an estimator
$\tauhat$ of $\tau$, the bias for group $\gg_m$ was defined as
$$
\bias(m) =
\EE\Bigl[\tauhat(\X)|\X\in\gg_m\Bigr] -
  \EE\Bigl[\tau(\X)|\X\in\gg_m\Bigr],
\hskip15pt m=1,\ldots,M.
$$ 
Recall that our simulation experiments were replicated independently $B$
times.  Let $\gg_{m,b}$ denote those $\x$ values that lie within the $q_m$
quantile of the propensity score from realization $b$.  Let $\tauhat_b$
be the ITE estimator from realization $b$.  The conditional
bias was estimated by
$$
\widehat{\bias}(m) =
\frac{1}{B}\sum_{b=1}^B\tauhat_{m,b}  - \frac{1}{B}\sum_{b=1}^B\tau_{m,b}\
$$
where
$$
\tauhat_{m,b} = \frac{1}{\# \gg_{m,b}}\sum_{\x_i\in\gg_{m,b}}  \tauhat_b(\x_i),\hskip10pt
\tau_{m,b} = \frac{1}{\# \gg_{m,b}}\sum_{\x_i\in\gg_{m,b}}  \tau(\x_i).
$$
Similarly, we define the conditional RMSE of $\tauhat$ by
$$
\text{RMSE} (m) = \sqrt{\EE\[\Bigl(\tauhat(\X)- \tau(\X)\Bigr)^2\,\biggl|\,\X\in\gg_m\]},
\hskip15pt m=1,\ldots,M,
$$
which we estimated using
$$
\widehat{\text{RMSE}}(m)
= \sqrt{\frac{1}{B}\sum_{b=1}^B 
\frac{1}{\# \gg_{m,b}}\sum_{\x_i\in\gg_{m,b}}  
\Bigl[\tauhat_b(\x_i) - \tau_b(\x_i)\Bigr]^2}.
$$

\subsection{Results}

\begin{figure}[phtb] 
\bct
\resizebox{2.9in}{!}{\includegraphics[page=1]{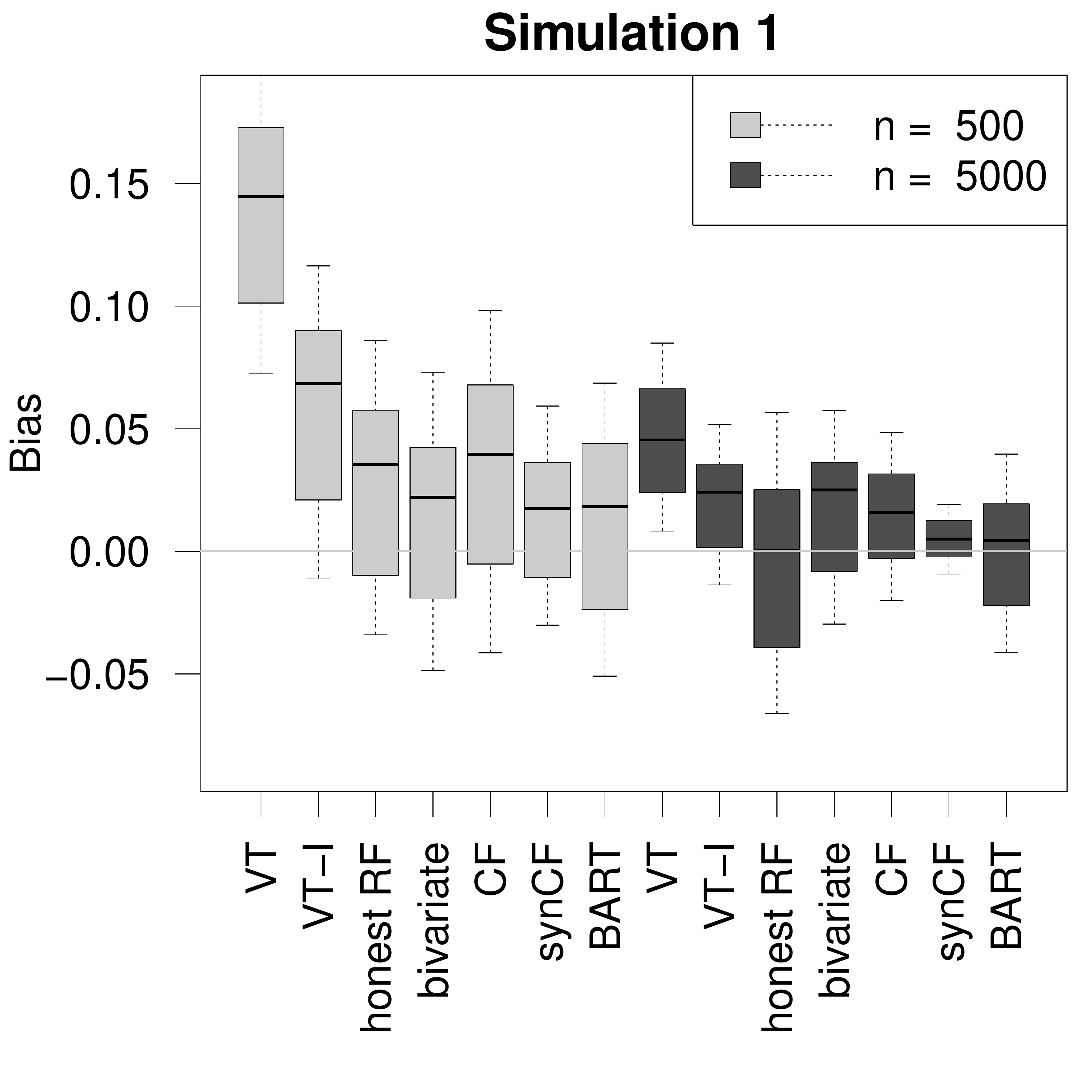}}
\resizebox{2.9in}{!}{\includegraphics[page=1]{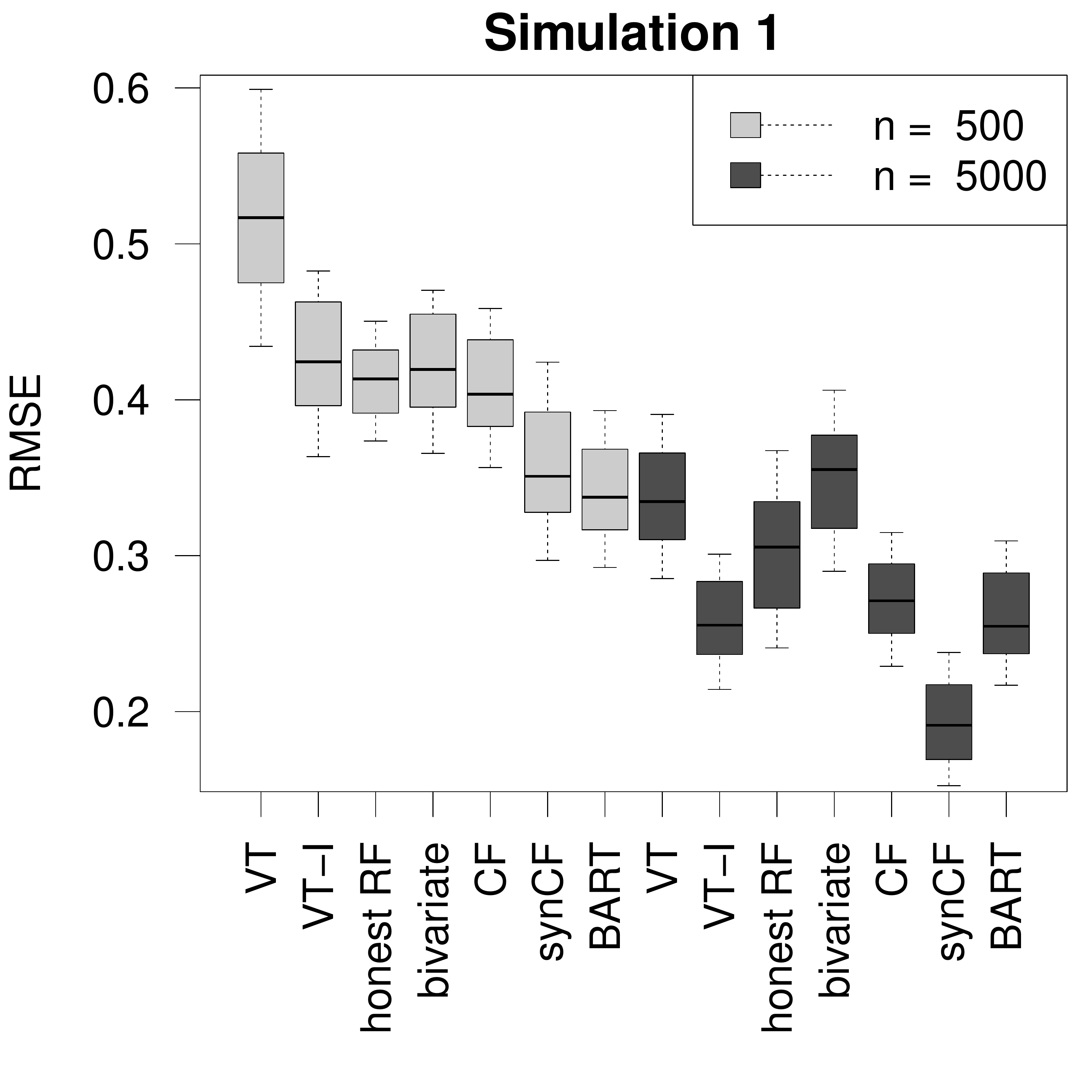}}
\resizebox{2.9in}{!}{\includegraphics[page=2]{benchmark_bias_summary.pdf}}
\resizebox{2.9in}{!}{\includegraphics[page=2]{benchmark_mse_summary.pdf}}
\resizebox{2.9in}{!}{\includegraphics[page=3]{benchmark_bias_summary.pdf}}
\resizebox{2.9in}{!}{\includegraphics[page=3]{benchmark_mse_summary.pdf}}
\caption{\em Bias and RMSE results from simulation experiments.}
\label{sim.results}
\ect
\end{figure}

Figure~\ref{sim.results} displays the conditional bias and RMSE for
each method for each of the simulation experiments.  Light and dark
boxplots display results for the small and larger sample sizes,
$n=500$ and $n=5000$; the left and right panels display bias and RMSE,
respectively.  Each boxplot displays $M$ values for the performance
measure evaluated at each of the $M$ stratified propensity score
groups.  We used a value of $M=100$ throughout.

\spacingset{1} 
\spacingset{1.45} 

Considering the RMSE results (right panels), it is clear that
counterfactual synthetic forests, $\tauhat_\SCF$, is generally the
best of all procedures, with results improving with increasing $n$.
The BART procedure, $\tauhat_\bart$, is comparable or slightly better
in simulations 1 and 2 when $n=500$, but $\tauhat_\SCF$ dominates when
the sample size increases to $n=5000$.  In simulation 3,
$\tauhat_\SCF$ is superior regardless of sample size. It is
interesting to observe that counterfactual forests, $\tauhat_\CF$,
which do not use synthetic forests for prediction, is systematically
worse than $\tauhat_\SCF$.  In fact, its performance is generally
about the same as $\tauhat_\VTI$ and the same as $\tauhat_\bart$ when
$n=5000$.  Regarding $\tauhat_\VTI$, it is interesting to observe how
it systematically outperforms $\tauhat_\VT$ regardless of simulation
or sample size.  This shows that augmenting the design matrix to
include treatment interactions really improves adaptivity of VT
forests.  Finally, the least successful procedure (for the large sample
size simulations) was the bivariate imputation method,
$\tauhat_{\biv}$.  Recall that the bivariate procedure differs from
the other procedures in that it uses mean imputation rather than
regression modeling of the outcome for ITE estimation.  This may
explain its poorer performance.  Following $\tauhat_{\biv}$ in terms
of overall performance, are $\tauhat_\VT$ and $\tauhat_\honest$, with
$\tauhat_\honest$ somewhere in between $\tauhat_\VT$ and $\tauhat_\VTI$.
This completes the discussion of the RMSE.  The results for bias (left
panels) generally mirror those for RMSE.  One interesting finding,
however, is the tightness of the range of bias values for $\tauhat_\SCF$
when $n=5000$.  This shows that with increasing sample size,
$\tauhat_\SCF$ gives consistently low bias even across extreme
propensity score values.

\section{Project Aware: a counterfactual approach to understanding
  the role of drug use in sexual risk}

Project Aware was a randomized clinical trial performed in nine
sexually transmitted disease clinics in the United States. The primary
aim was to test whether brief risk-reduction counseling performed at
the time of an HIV test had any impact on subsequent incidence of
sexually transmitted infections (STIs). The results showed no impact
of risk-reduction counseling on STIs. Neither were there any substance
use interactions of the impact of risk-reduction counseling;
however, substance use was associated with higher levels of STIs at
follow-up. Other research has shown that substance use is associated
with higher rates of HIV testing, and Black women showing the highest
rates of HIV testing in substance use treatment
clinics~\citep{hernandez2016self}. Since substance use is associated
with risky sexual activity, detecting the dynamics of this
relationship can contribute to preventive and educational efforts to
control the spread of HIV. Our procedures for causal analysis of
heterogeneity of effects in observational data should equalize the
observed characteristics among substance use and non-substance use
participants, 
thereby removing any impact of background imbalance in factors that
may be related to relationship of substance use on sexual risk. Our
procedure then allows an exploration of background factors that are
truly related to this causal effect, conditional on all confounding
factors being in the feature set.

To explore this issue of how substance use plays a role in sexual
risk, we pursued an analysis in which the treatment (exposure)
variable $T$ was defined as drug use status of an individual (0 $=$ no
substance use in the prior 6 months, 1 $=$ any substance use in the
prior 6 months leading to the study).
For our outcome, we used number of unprotected sex
acts within the last six months as reported by the
individual. Although Project Aware was randomized on the primary
outcome (risk-reduction counseling), analysis of secondary outcomes
such as substance use should be treated as if from an observational
study. Indeed, unbalancedness of the data for drug use can be
gleaned from Table~\ref{rlcf} which displays results from a logistic
regression in which drug use status was used for the dependent
variable ($n=2813$, $p=99$). The list of significant variables
suggests the data is unbalanced and indicates that inferential methods
should be considered carefully. Thus Table~\ref{rlreg}, which displays
the results from a linear regression using number of unprotected sex
acts as the dependent variable, should be interpreted with caution.
Table~\ref{rlreg} suggests there is no overall exposure effect of drug
use, although several variables have significant drug-interactions.

\spacingset{1.0} 
\begin{table}[phtb]
  \centering
  \caption{\em Difference in variables by drug use illustrating
    unbalancedness of Aware data. Only significant variables (p-value
    $<$ 0.05) from logistic regression analysis are displayed for
    clarity.}
  \label{rlcf}
  \begin{tabular}{@{}lcccc@{}}
    \specialrule{.15em}{.1em}{.1em}
			& Estimate & Std. Error & Z     & p-value \\ \midrule
Race & -0.28 & 0.11 & -2.50 & 0.01 \\ 
Chlamydia & 0.34 & 0.15 & 2.30 & 0.02 \\ 
Site 2 & -0.62 & 0.16 & -3.94 & 0.00 \\ 
Site 4 & -0.53 & 0.16 & -3.23 & 0.00 \\ 
Site 6 & 0.44 & 0.18 & 2.43 & 0.01 \\ 
Site 7 & -0.65 & 0.15 & -4.22 & 0.00 \\ 
Site 8 & 0.95 & 0.21 & 4.51 & 0.00 \\ 
HIV risk & 0.17 & 0.03 & 5.14 & 0.00 \\ 
CESD & 0.02 & 0.01 & 3.13 & 0.00 \\ 
Condom change 2 & -0.24 & 0.12 & -2.07 & 0.04 \\ 
Marriage & 0.08 & 0.03 & 2.76 & 0.01 \\ 
In Jail ever & 0.42 & 0.10 & 4.07 & 0.00 \\ 
AA/NA last 6 months 1& 0.69 & 0.23 & 3.04 & 0.00 \\ 
Frequency of injection & 0.18 & 0.07 & 2.49 & 0.01 \\ 
Gender & -0.39 & 0.10 & -4.00 & 0.00 \\ 
\specialrule{.15em}{.1em}{.1em}
  \end{tabular}
\end{table}

\begin{table}[phtb]
  \centering
  \caption{\em Linear regression where dependent variable is number of
    unprotected sex acts from Aware data. Only variables with p-value
    $<$ 0.10 from regression analysis are displayed for clarity.}
  \label{rlreg}
  \begin{tabular}{@{}lcccc@{}}
    \specialrule{.15em}{.1em}{.1em}
& Estimate & Std. Error & Z     & p-value \\ \midrule
Intercept & -5.06 & 22.17 & -0.23 & 0.82 \\ 
Drug & 9.59 & 29.72 & 0.32 & 0.75 \\ 
HCV2 & 8.14 & 4.09 & 1.99 & 0.05 \\ 
Site 2 & -14.00 & 6.82 & -2.05 & 0.04 \\ 
HIV risk & 3.89 & 1.41 & 2.76 & 0.01 \\ 
Condom change 3 & -17.23 & 6.52 & -2.64 & 0.01 \\ 
Condom change 5 & -21.98 & 6.65 & -3.30 & 0.00 \\ 
Visit opthamologist & -16.46 & 8.17 & -2.01 & 0.04 \\ 
Number visit opthamologist & 9.13 & 3.93 & 2.33 & 0.02 \\ 
Marriage & -2.13 & 1.17 & -1.81 & 0.07 \\ 
Smoke & 53.13 & 16.26 & 3.27 & 0.00 \\ 
Number cigarette per day & -14.63 & 4.76 & -3.07 & 0.00 \\ 
Drug x CESD & 0.88 & 0.47 & 1.88 & 0.06 \\ 
Drug x Condom change 2 & -24.70 & 6.83 & -3.62 & 0.00 \\ 
Drug x Condom change 3 & -25.82 & 8.75 & -2.95 & 0.00 \\ 
Drug x Condom change 4 & -37.58 & 14.20 & -2.65 & 0.01 \\ 
Drug x Condom change 5 & -28.78 & 9.33 & -3.08 & 0.00 \\ 
Drug x Visit dentist & -16.36 & 8.26 & -1.98 & 0.05 \\ 
Drug x Smoke & -34.71 & 20.53 & -1.69 & 0.09 \\ 
Drug x Number cigarette per day & 9.81 & 5.95 & 1.65 & 0.10 \\ 
\specialrule{.15em}{.1em}{.1em}
  \end{tabular}
\end{table}
\spacingset{1.45}

\begin{figure}[phtb!] 	
  \bct
\hskip-40pt
  \resizebox{6.85in}{!}{\includegraphics[page=1]{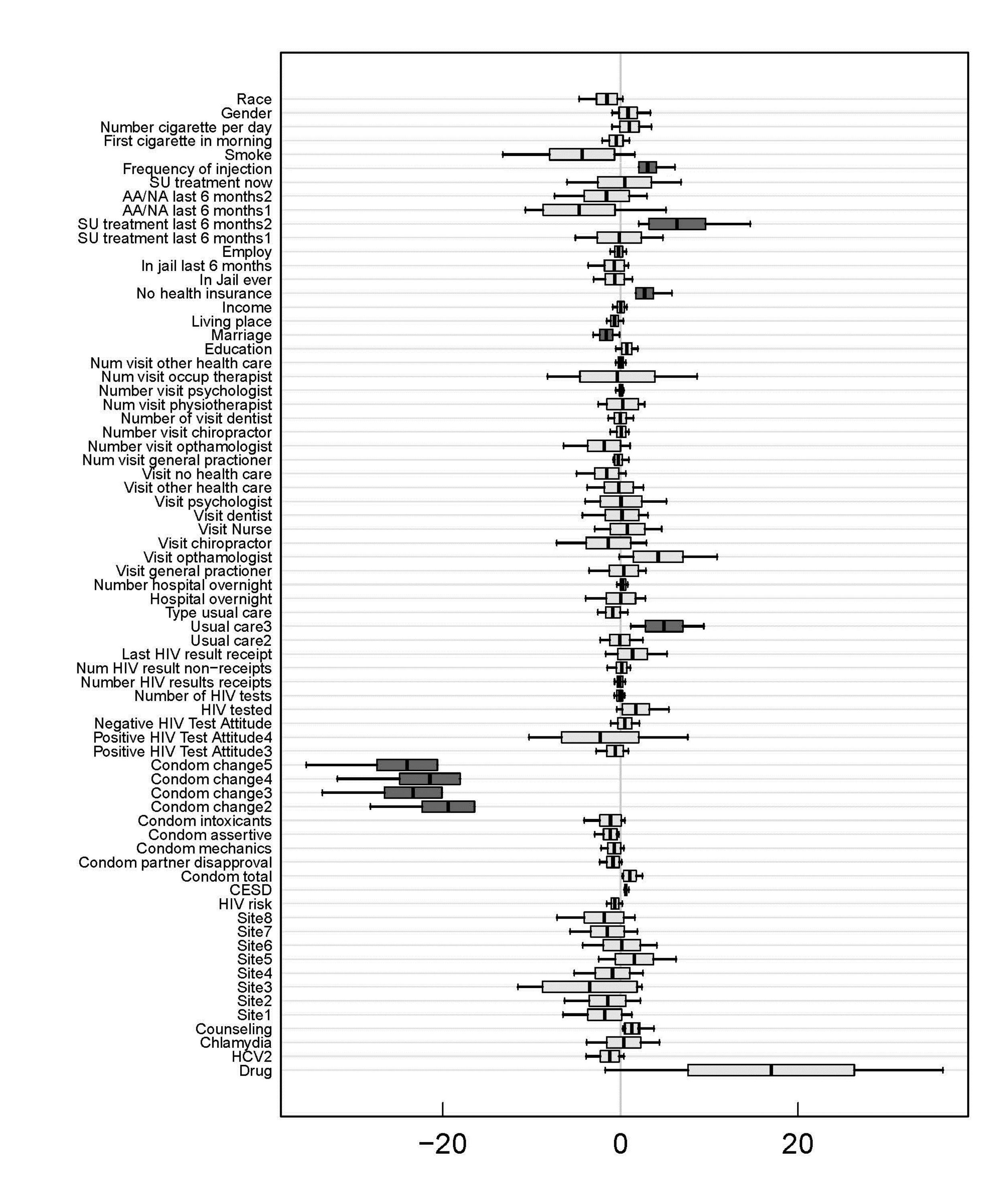}}
  \caption{\em Confidence intervals for all coefficients of linear model
    used in Table~3.   Intervals determined using subsampling. Dark colored boxplots indicate
    variables with p-value $<$ .05.}\label{rlcof}
\ect
\end{figure}

However, in order to avoid drawing potentially flawed conclusions from
an analysis like Table~\ref{rlreg}, we applied our counterfactual
synthetic approach, $\tauhat_\SCF$.
A synthetic forest was fit separately to each exposure group using
number of unprotected sex acts as the dependent variable. This yielded
estimated causal effects $\{\tauhat_\SCF(\x_i),i=1,\ldots,n\}$ for
$\tau(\x)$ defined as the mean difference in number of unprotected sex
acts for drug versus non-drug users.  The estimated causal effects
were then used as dependent variables in a linear regression
analysis. This is convenient because the estimated coefficients from
the regression analysis can be interpreted in terms of subgroup causal
differences (we elaborate on this point shortly). In order to derive
valid standard errors and confidence regions for the estimated
coefficients, the entire procedure was subsampled. That is, we drew a
sample of size $m$ without replacement. The subsampled data was then
fit using synthetic forests as described above, and the resulting
estimated causal effects used as the dependent variable in a linear
regression. The procedure was repeated 1000 times independently. A
subsampling size of $m = n/10$ was used. The confidence regions of the
resulting coefficients are displayed in Figure~\ref{rlcof}.  Table~\ref{rlrs}
displays the coefficients for significant values (p-values $<.05$).
We note that bootstrapping could have have been used as another means
to generate nonparametric p-values and confidence regions.  However,
we prefer subsampling because of its computational speed and general
robustness~\citep{politis1999subsampling}.

\spacingset{1} 
\spacingset{1.45} 

To interpret the coefficients in Table~\ref{rlrs}, it is useful to write the
true model for the outcome (number of unprotected sex acts)
as $Y=f(\X,T)+\eps$, where
$$
f(\X,T) =  \a_0 T + h(\X,T),
$$
and $h$ is some unknown function.  Under the
assumption of SITA, and using the same calculations
as~\mref{regression.validation}, we have
$$
\tau(\x) 
= f(\x,1)-f(\x,0)
=\a_0 + h(\x,1) - h(\x,0).
$$
Now since we assume a linear model $\a + \sum_{j=1}^p
\beta_j x_j$ for the ITE, we have
$$
\a_0 + h(\x,1) - h(\x,0) =  \a + \sum_{j=1}^p \beta_j x_j.
$$ From this we can infer that the intercept in Table~\ref{rlrs} is an
overall measure of the exposure effect of drug use, $\a_0$ (this is
why the intercept term is listed as drug use). Here the estimated
coefficient is 17.0. The positive coefficient implies that on average
drug users have significantly more unprotected sex acts than
non-drug users (significance here is slightly larger than 5\%).

\vskip10pt
\spacingset{1} 
	\begin{table}[phtb!]
		\centering
		\caption{\em Linear regression of Aware data with
                  dependent variable equal to the estimated causal
                  effects $\{\tauhat_\SCF(\x_i),i=1,\ldots,n\}$ from
                  counterfactual synthetic random forests.
                  Causal effect is defined as the mean
                  difference in unprotected sex acts for drug users
                  versus non-drug users. Standard
                  errors and significance of linear model coefficients
                  were determined using subsampling.  For clarity, only
                  significant variables with p-value $<$ 0.05 are
                  displayed (the intercept is provided for reference
                  but is not significant).}
		\label{rlrs}
		\begin{tabular}{@{}lccc@{}}
			\specialrule{.15em}{.1em}{.1em}
			& Estimate & Std. Error & Z     \\ \midrule
			Intercept (drug use) & 16.97 & 9.36 & 1.81 \\ 
			CESD & 0.60 & 0.13 & 4.54 \\ 
			Condom change 2 & -19.38 & 2.96 & -6.56 \\ 
			Condom change 3 & -23.33 & 3.23 & -7.22 \\ 
			Condom change 4 & -21.46 & 3.39 & -6.32 \\ 
			Condom change 5 & -24.02 & 3.41 & -7.04 \\ 
			Usual care 3 & 4.91 & 2.11 & 2.33 \\ 
			Marriage & -1.61 & 0.73 & -2.21 \\ 
			No health insurance & 2.72 & 0.99 & 2.75 \\ 
			SU treatment last 6 months 2 & 6.38 & 3.20 & 2.00 \\ 
			Frequency of injection & 3.59 & 1.77 & 2.02 \\ 
			\specialrule{.15em}{.1em}{.1em}
		\end{tabular}
	\end{table}
\spacingset{1.45} 

The remaining coefficients in Table~\ref{rlrs} describe how the effect
of drug use on sexual risk is modulated by other factors.  Under our
linear model, we have
$$
h(\x, 1) -h(\x, 0)=\sum_{j=1}^{p}\b_jx_j.
$$ 
Because $h(\x, 1) - h(\x, 0)$ represents how much a subgroup
deviates from the overall causal effect, each coefficient in Table
\ref{rlrs} quantifies the effect of a specific subgroup on drug use
differences. Consider for example, the variable ``Frequency of
injection" which is a continuous variable representing frequency of
injections in drug users. Because its estimated coefficient is 3.6,
this means the difference in unprotected sex acts between drug and and
non-drug users, which is positive, becomes even wider for high
frequency drug users. Another risky factor is ``No health
insurance", which is an indicator of lack of health insurance
coverage. Because its estimated coefficient is 2.7, we can take this
to mean that the increase in sexual risk for an individual without
health insurance is more pronounced in drug users.  As another
example, consider the variable ``Condom change" which is an ordinal
categorical variable measuring an individual's stage of change
with respect to condom use behavior. The baseline level is a
``precontemplator", who is an individual who has not envisioned using
condoms. The second level ``contemplator" is an individual
contemplating using condoms. Further increasing levels measure even
more willingness to utilize condoms. All coefficients for Condom
change in Table \ref{rlrs} are negative, and therefore if an
individual is more willing to utilize safe condom practice (relative
to the baseline condition), the difference in number of unprotected
sex acts diminishes between drug and non-drug users. Other variables
that have a subgroup effect are Marriage (whether an individual is
married), CESD (Center for Epidemiological Studies Depression Scale),
and SU treatment last 6 months (substance abuse 
treatment in last 6 months). In all of these, the pattern is similar
to before. With more risky behavior (with depression) the number of
unprotected sex acts increases for non-drug users relative to
drug users, but as risky behavior decreases (e.g.\ married), the
effect of drug use diminishes.

\begin{figure}[phtb!]
\centering
\includegraphics[width=\linewidth]{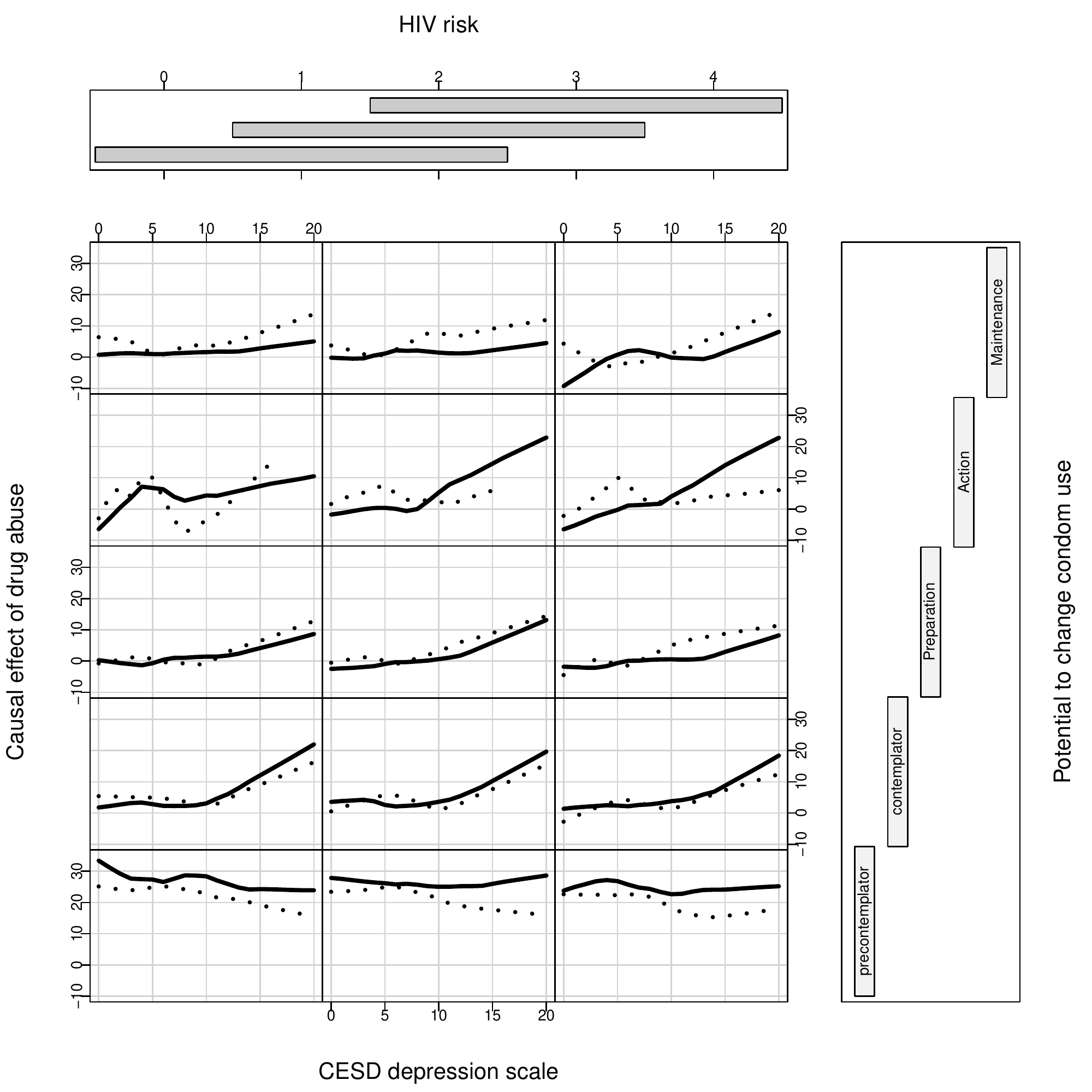}
\vskip-30pt
\includegraphics[width=\linewidth]{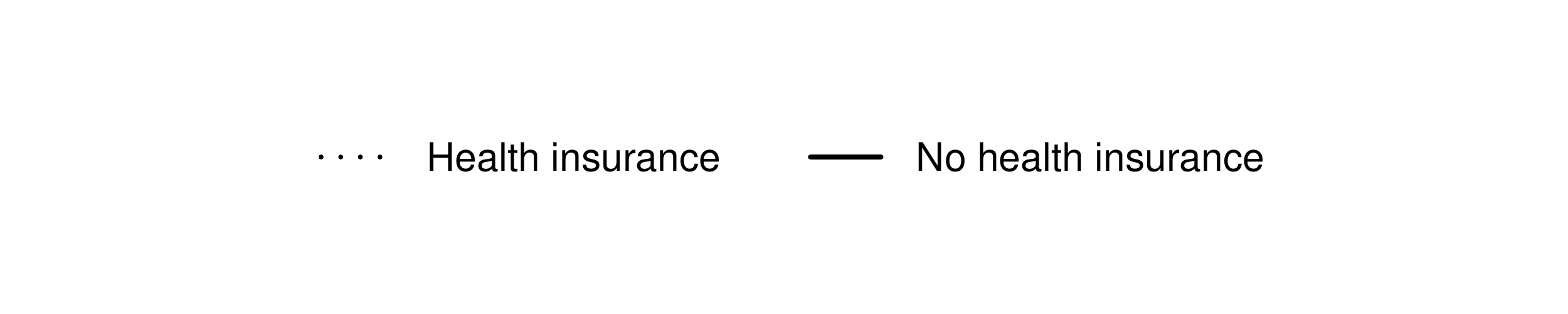}	
\vskip-20pt
\caption{\em RF estimated causal effect of drug use plotted
  against  CESD depression for individuals with and without health
insurance.  Values are conditioned on Condom change (vertical
conditional axis) and HIV risk (horizontal conditional axis).}\label{coplot}
\end{figure}

Figure~\ref{coplot} displays a coplot of the RF estimated causal
effects $\{\tauhat_\SCF(\x_i),i=1,\ldots,n\}$ as a function of several
variables.  The coplot is another useful tool that can be used to
explore causal relationships.  We use it to uncover relationships that
may be hidden in the linear regression analysis.  The RF causal
effects are plotted against CESD depression for individuals with and
without health insurance.  Conditioning is on the variables Condom
change (vertical conditioning) and HIV risk (horizontal conditioning).
HIV risk a self-rated variable and of potential importance and was
included even though it was not significant in the linear regression
analysis.  For patients with potential to change condom use (rows 2
through 5), increased depression levels leads to an increased causal
effect of drug use, which is slightly accentuated for high HIV risk
(plots going from left to right).  The effect of health insurance is
however minimal.  On the other hand, for individuals with low
potential to change condom use (bottom row), the estimated exposure
effect is generally high, regardless of depression, but is reduced if
the individual has health insurance.

\spacingset{1} 
\spacingset{1.45}

\section{Discussion}

In observational data with complex heterogeneity of treatment effect,
individual estimates of treatment effect can be obtained in a
principled way by directly modeling the response outcome.  However,
successful estimation mandates highly adaptive and accurate regression
methodology and for this we relied on RF, a machine learning method
with well known properties for accurate estimation in complex
nonparametric regression settings.  However, care must be used when
applying RF for casual inference.  We encourage the use of out-of-bag
estimation, a simple but underappreciated out-of-sample technique for
improving accuracy.  We also recommend that when selecting a RF
approach, that it should have some means for encouraging adaptivity to
confounding, i.e.\ that it can accurately model potentially separate
regression surfaces for each of the treatment groups.  One example of
this is the extension to VT, which expands the design matrix to
include all pairwise interactions of variables with the treatment, a
method we call $\tauhat_\VTI$, and described in the paper
by~\cite{foster2011subgroup}.  We found that this simple extension,
when coupled with out-of-bagging, significantly improved performance
of VT.  Another promising method was counterfactual synthetic forests
$\tauhat_\SCF$, which generally had the best performance among all
methods, and was superior in the larger sample size simulations,
outperforming even the highly adaptive BART method.  The larger sample
size requirement is not so surprising as having to grow separate
forests causes some loss of efficiency; this being however mitigated
by its superior bias properties which take hold with increasing $n$.

In looking back, we can now see that the success of counterfactual
synthetic RF can be attributed to three separate effects: (a) fitting
separate forests to each treatment group, which improves adaptivity to
confounding; (b) replacing Breiman forests with synthetic forests,
which reduces bias; and (c) utilizing OOB estimation, which improves
accuracy.  Computationally, counterfactual synthetic RF are easily
implemented with available software and have the added attraction that
they reduce parameter tuning.  The latter is a
consequence and advantage of using synthetic forests.  A synthetic
forest is constructed using RF base learners, each of these being
constructed under different nodesize and mtry tuning parameters.
Correctly specifying mtry and nodesize is important for good
performance in Breiman forests.  The optimal value will depend on
whether the setting is large $n$, large $p$, or large $p$ and large
$n$.  With synthetic forests this problem is alleviated by building RF
base learners under different tuning parameter values.

Importantly, and underlying all of this, is the potential outcomes
model, a powerful hypothetical approach to causation.  The challenge
is being able to properly fit the potential outcomes model and for
this, as discussed above, we relied on the sophisticated machinery of
RF.  We emphasize that the direct approach of the potential outcomes
model is well suited for personalized inference via the ITE.
Estimated ITE values from RF can be readily analyzed using standard
regression models to yield direct inferential statements for not only
overall treatment effect, but also interactions, thus facilitating
inference beyond the traditional ATE population-centric viewpoint.
Using the Aware data we showed how counterfactual ITE estimates from
counterfactual synthetic forests could be explored to understand
causal relations.  This revealed interesting connections between
risky behavior, drug use, and sexual risk.  The analysis corrects
for any observed differences by the exposure variable, so to the
extent that we have observed the important confounding variables, this
result can tentatively be considered causal, though caution should be
used due to this assumption.  Clearly, this type of analysis, which
controls for observed confounding gives additional and important
insights above simple observed drug usage differences.  We also note
that although we used linear regression for interpretation in
this analysis, it is possible to utilize other methods as well.  The
counterfactual synthetic forest procedure provides a pipleline that can be
connected with many types of analyses, such as the conditional plots
that were also used in the Aware data analysis.

\section*{Acknowledgments}

This work was supported by the National Institutes of Health
[R01CA16373 to H.I. and U.B.K., R21DA038641 to D.J.F.]
and by the Patient Centered Outcomes
Research [ME-1403-12907 to D.J.F, H.I., M.L and S.S.].


\end{document}